\documentclass[10pt,twocolumn,letterpaper]{article}

\usepackage{cvpr}      

\usepackage{graphicx}
\usepackage{amsmath}
\usepackage{booktabs}
\usepackage{amsfonts}
\usepackage{multirow}
\usepackage{utfsym}
\usepackage{ulem}
\usepackage{float}

\usepackage[pagebackref,breaklinks,colorlinks]{hyperref}

\usepackage[capitalize]{cleveref}
\crefname{section}{Sec.}{Secs.}
\Crefname{section}{Section}{Sections}
\Crefname{table}{Table}{Tables}
\crefname{table}{Tab.}{Tabs.}

\def\cvprPaperID{*****} 
\def\confName{CVPR}
\def\confYear{2022}

\begin{document}

\title{The Solution for the sequential task continual learning track of the 2nd Greater Bay Area International Algorithm Competition}

\author{Sishun Pan,
Xixian Wu,
Tingmin Li,
Longfei Huang,
Mingxu Feng,
Zhonghua Wan,
Yang Yang\thanks{Corresponding author: 	yyang@njust.edu.cn} 
\\Nanjing University of Science and Technology}

\maketitle

\begin{abstract}
This paper presents a data-free, parameter-isolation-based continual learning algorithm we developed for the sequential task continual learning track of the 2nd Greater Bay Area International Algorithm Competition. The method learns an independent parameter subspace for each task within the network's convolutional and linear layers and freezes the batch normalization layers after the first task. Specifically, for domain incremental setting where all domains share a classification head, we freeze the shared classification head after first task is completed, effectively solving the issue of catastrophic forgetting. Additionally, facing the challenge of domain incremental settings without providing a task identity, we designed an inference task identity strategy, selecting an appropriate mask matrix for each sample. Furthermore, we introduced a gradient supplementation strategy to enhance the importance of unselected parameters for the current task, facilitating learning for new tasks. We also implemented an adaptive importance scoring strategy that dynamically adjusts the  amount of parameters to optimize single-task performance while reducing parameter usage. Moreover, considering the limitations of storage space and inference time, we designed a mask matrix compression strategy to save storage space and improve the speed of encryption and decryption of the mask matrix. Our approach does not require expanding the core network or using external auxiliary networks or data, and performs well under both task incremental and domain incremental settings. This solution ultimately won a second-place prize in the competition.

\end{abstract}

\section{Introduction}
\label{sec:intro}

\begin{figure}[t]
	\centering
	\includegraphics[width=\linewidth]{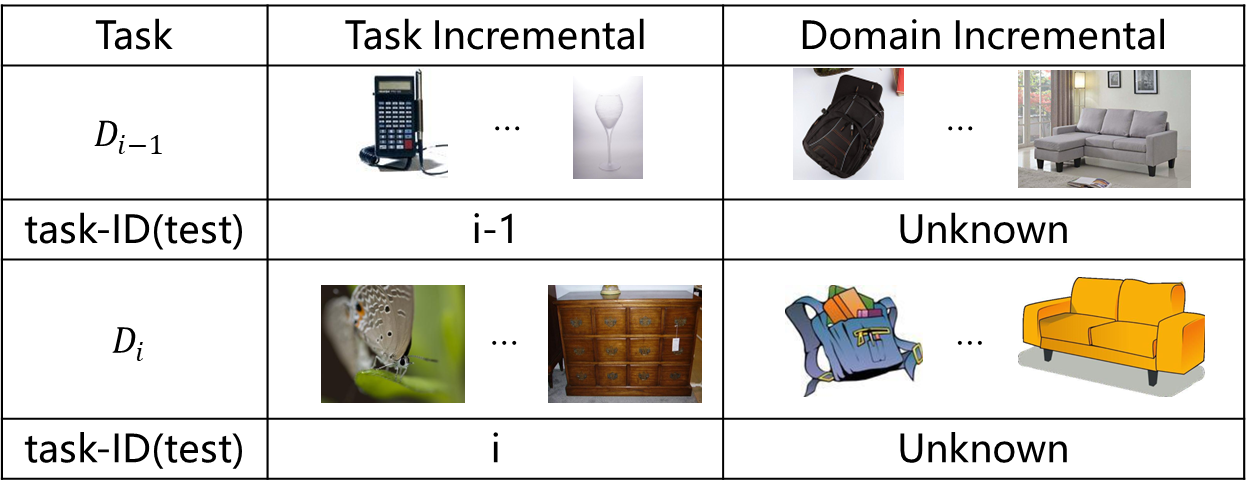}
	\caption{Examples of different incremental learning tasks. The left column shows a sequence of tasks in the Task Incremental setting, including examples for the test of historical task \(D_{i-1}\) and the current task \(D_i\). The right column displays examples from various domains in the Domain Incremental setting, where the task ID is unknown during testing.}\label{fig:task_introduction}
\end{figure}
In recent years, deep neural networks have excelled in single-task scenarios\cite{yang2024robust}\cite{yang2024alignment}\cite{yang2023contextualized}\cite{yang2021corporate}. However, in dynamic environments where data arrives incrementally, deep models need to dynamically update with new data to learn new knowledge\cite{yang2024not}. Therefore, continual Learning has garnered widespread attention. Continual learning\cite{yang2021learning} aims to enable models to progressively learn a series of tasks, acquiring new knowledge on new tasks while maximizing the retention of knowledge from historical tasks. In this competition, we faced the challenge of designing an algorithm that is applicable to both task incremental and domain incremental settings, as shown in the Figure \ref{fig:task_introduction}. These two settings differ fundamentally: in task incremental settings, each task uses an independent classification head, while in domain incremental settings, all domains share a single classification head, and no task identity is provided during the testing phase of domain incremental settings\cite{zhou2021learning}\cite{zhang2021harmi}. Additionally, the competition rules prohibit saving historical task training data, external auxiliary data, network parameters, and expanding the core neural network.

To address these challenges, we designed a data-free, parameter-isolation-based continual learning method, which encompasses the following five aspects: (1) We use parameter isolation as the primary strategy, learning independent parameter subspaces for each task within the linear and convolutional layers. Since the parameters of the batch normalization layer dynamically change with the observed data, we freeze the batch normalization layer after the first task. To adapt the algorithm for domain incremental settings, we also freeze the classification head after the first task to fundamentally prevent catastrophic forgetting. (2) We designed a strategy to infer the task identity for each sample, addressing the issue of no task identity being provided during the inference phase of domain incremental settings. (3) We introduced a gradient supplementation strategy to facilitate the learning of new tasks, thereby enhancing the accuracy of each task. (4) We implemented a strategy to dynamically adjust the number of parameters, reducing parameter usage for each task, making the method more suitable for long task sequences. (5) We proposed a mask matrix compression strategy that not only saves storage space for the mask matrix but also speeds up the encryption and decryption processes of the mask matrix.

Our approach ranked third in the final stage and won a second-place prize in the competition. In the remainder of this technical report, we will introduce the detailed architecture of our solution for this challenge.


\section{Related Work}
\label{sec:Related}
Continual learning\cite{wang2024comprehensive} refers to a model continuously learning a series of tasks while retaining knowledge from previous tasks. Continual learning approaches can be broadly divided into
five categories.

\textbf{Regularization-based Methods} add explicit regularization terms to balance the knowledge between new and old tasks. Kirkpatrick et al.\cite{ewc} develop an algorithm called elastic weight consolidation (EWC), which slows down learning on certain weights based on their importance to previously seen tasks. Wang et al.\cite{afec} propose a novel approach named Active Forgetting with synaptic Expansion-Convergence (AFEC), which dynamically expands parameters to learn each new task and then selectively combines them. Lin et al.\cite{connector} propose to employ mode connectivity in loss landscapes to achieve better plasticity-stability trade-off without any previous samples.

\textbf{Replay-based Methods} store samples from old tasks in memory or use additional generative models to replay historical data. Buzzega et al.\cite{der} propose a novel continual learning baseline called Dark Experience Replay (DER), which improves on Experience Replay (ER) by relying on dark knowledge to distill past experiences. 
Wang et al.\cite{replay_2} adopted a generative model to produce additional (imaginary) data based on limited memory, effectively reducing catastrophic forgetting.

\textbf{Optimization-based methods} use specially designed optimization processes to reduce forgetting, such as ensuring that the gradient directions of model updates are orthogonal to the feature space of historical task data. Farajtabar et al.\cite{OGD} introduce the Orthogonal Gradient Descent (OGD) method, which achieves this by projecting gradients from new tasks onto a subspace where the neural network's output on previous tasks remains unchanged, while ensuring the projected gradient remains beneficial for learning the new task. Wang et al.\cite{Adam-NSCL} proposed a new network training algorithm called Adam NSCL, which sequentially optimizes network parameters in the null space of previous tasks, balancing the network's plasticity and stability.

\textbf{Representation-based methods} aim to prevent catastrophic forgetting by leveraging the advantages of learned representations. Fini et al. \cite{CaSSLe} designed a framework for continuous self-supervised visual representation learning, significantly enhancing the quality of the learned representations. Cha et al. \cite{Co2L} learn representations using the contrastive learning objective and preserve these representations through a self-supervised distillation step, thereby ensuring the maintenance of transferable representations.

\textbf{Architecture-based methods} explicitly address catastrophic forgetting by constructing task-specific parameters. Serra et al.\cite{hat} propose a task-based hard attention mechanism that preserves previous tasks' information without affecting the current task's learning. Yang et al.\cite{yang2019adaptive} develop an incremental adaptive deep model (IADM) that allows the deep model structure to scale with streaming data. Yang et al. \cite{yang2021cost} address the challenges of capacity requirements, scalability, and sustainability in incremental deep models through their design of the Cost-Effective Incremental Deep Model, which slows down the rate of forgetting. Kang et al.\cite{wsn} propose a continual learning method referred to as Winning SubNetworks (WSN), which sequentially learns and selects an optimal subnetwork for each task, inherently immune to catastrophic forgetting since each selected subnetwork model does not interfere with others.

\section{Methodology}
\subsection{Overall Architecture}


\begin{figure}[t]
\centering
\begin{subfigure}{0.5\textwidth}
    \centering
    \includegraphics[width=\textwidth]{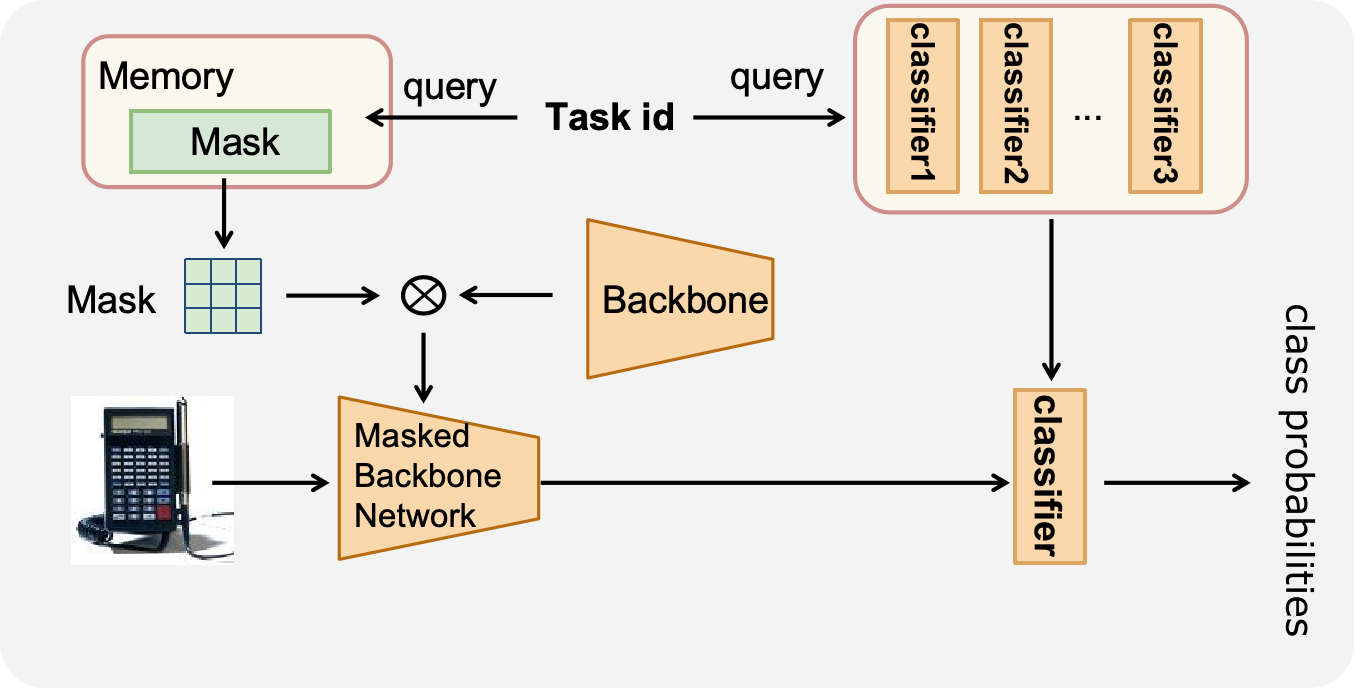}
    \caption{Task increment}
    \label{fig:a}
\end{subfigure}
\hfill 
\begin{subfigure}{0.5\textwidth}
    \centering
    \includegraphics[width=\textwidth]{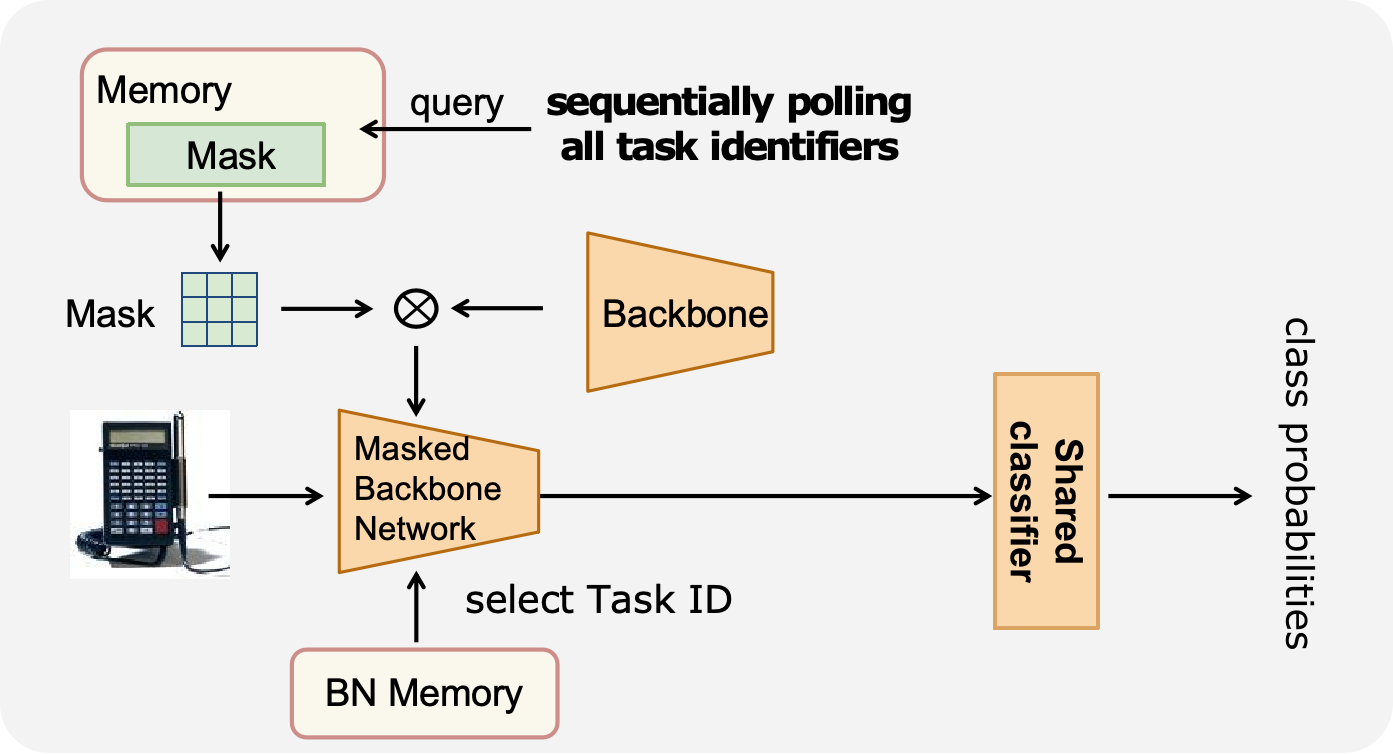}
    \caption{Domain increment}
    \label{fig:b}
\end{subfigure}
\caption{Overview of the system architecture. (a) We train the feature extractor and the task classifier \( k \) at task \( k \). (b) Transformer and adapter module with specific masks.}
\label{fig:Architecture}
\end{figure}

Figure \ref{fig:Architecture} illustrates our overall approach, which includes different learning and inference strategies for task increment and domain increment conditions. Figure 1(a) shows the overall method under task increment conditions, while Figure 1(b) displays the method under domain increment conditions. 

\subsection{Model Architecture}
Our scheme is based on a parameter isolation method called WSN\cite{wsn}. There are two main reasons for choosing WSN: First, as a parameter isolation method, WSN fundamentally eliminates catastrophic forgetting by assigning a subnetwork to each task. Second, WSN can selectively reuse parameters from previous tasks, effectively leveraging cross-task knowledge. This not only aids in knowledge transfer and generalization but also significantly reduces training convergence time. However, in the ResNet network provided for the competition, catastrophic forgetting may still occur due to the characteristics of the BN layers. During continual learning, the BN layers update with samples from different tasks, causing a mismatch with the frozen subnetworks. To address this, we made the following two improvements to WSN: (1) Under task increment conditions, after learning the first task, all BN layers are frozen, and subsequent tasks use the BN layers from the first task to prevent forgetting caused by changes in BN layers; (2) Under domain increment conditions, in addition to freezing the BN layers, since all domains share a classification head, we also freeze the final fully connected layer. These innovative changes make WSN better suited to this competition's challenges, effectively preventing forgetting.

\subsection{Domain Increment Inference Task ID}

In this competition, our approach must be suitable not only for task increment continual learning but also for domain increment continual learning. Unfortunately, WSN and other parameter isolation methods are not well-suited for domain increment learning. Under domain increment continual learning, during the testing phase, the model cannot access the current Task ID, and therefore cannot retrieve the corresponding Mask matrix for model inference. To solve this, we infer the Task ID to obtain the relevant Mask matrix. We store the mean and variance of all samples after the first convolution layer for each task. During inference, we calculate the difference in mean and variance after the first convolution layer under different Masks compared to the stored task mean and variance; the Task ID with the smallest difference indicates the Task ID for that sample. This method performs excellently in terms of storage and inference efficiency, only storing the mean and variance after the first layer of each task with minimal storage overhead. During ID inference, only the first layer of the model is required, ensuring extremely low inference overhead.

\subsection{Gradient Supplementation}

We have identified certain deficiencies with the WSN method. The importance matrix for parameters of the current task is initialized based on the importance matrix from previous tasks, tending to use parameters from previous tasks while neglecting others. Therefore, we have adopted a gradient supplementation strategy where each weight Theta is associated with an importance score s, updated as follows:
\[
\mathbf{s} \leftarrow \mathbf{s} - \begin{cases}
\eta \left( \frac{\partial \mathcal{L}}{\partial \mathbf{s}} \right) \cdot \gamma & \text{if } \mathbf{M}_{t-1} = 1 \text{ or } \mathbf{m}_t = 0 \\
\eta \left( \frac{\partial \mathcal{L}}{\partial \mathbf{s}} \right) & \text{otherwise}
\end{cases}
\]
, where \(\eta\) is the learning rate, \(\gamma\) is the gradient supplementation coefficient, \(\mathbf{M}_{t-1}\) is the historical task mask, and \(\mathbf{m}_t\) is the current task mask. In this way, the gradient of parameters not selected for the current task is supplemented, which is more beneficial for learning the current task.

\subsection{Dynamic Adjustment of Subnetwork Size}

In this competition, we face the challenge of long sequence task increment learning. The preliminary round involves incremental learning of 10 tasks, while the final round expands to 20 tasks. Given such a long sequence of task increments, it is crucial to appropriately allocate the size of each subnetwork to ensure the model still has sufficient learnable parameters for subsequent tasks. In this context, the traditional WSN does not consider this requirement. To address this, we introduce a dynamic threshold method, selecting network parameters that exceed a specific importance threshold. We independently set an adaptive threshold for each layer of the network, selecting weights in each layer that exceed $\alpha$ times the maximum importance of that layer. This method ensures that only the most critical network weights of the current task are utilized, leaving more learnable space for subsequent tasks and effectively securing the model's learning capability for future tasks. The importance scores for the parameters in the \( l \)-th layer are represented by the vector \( s^{(l)} \). We calculate a dynamic threshold \( \theta^{(l)} \) for each layer as:
\[
\theta^{(l)} = \alpha^{(l)} \cdot \max(s^{(l)})
\]
The masking vector \( m^{(l)} \) for each parameter in the \( l \)-th layer is defined as:
\[
m^{(l)}_i = \begin{cases} 
1 & \text{if } s^{(l)}_i \geq \theta^{(l)} \\
0 & \text{otherwise}
\end{cases}
\]
The masked weights of the model in the \( l \)-th layer, \( \mathbf{w}'^{(l)} \), are then calculated by the Hadamard product (denoted by \( \odot \)):
\[
\mathbf{w}'^{(l)} = \mathbf{w}^{(l)} \odot \mathbf{m}^{(l)}
\]
This modification ensures that each layer individually activates its parameters based on their respective importance scores, enhancing layer-specific optimization and efficiency in learning.

\subsection{Mask Matrix Compression}

Although the Mask matrix consists of 0 and 1, storing them as floats still occupies substantial storage space. In WSN, Huffman coding was proposed for compression. However, in our actual use, Huffman coding not only provided an inadequate compression ratio but also involved significant computational overhead. We propose a simple and efficient compression method: using each bit of a 32-bit integer to store a task's Mask matrix. With \( T \) tasks, corresponding to \( T \) mask matrices \( M_0, M_1, \ldots, M_{T-1} \), each element \( M_k[i] \) of a mask matrix \( M_k \) represents the value at position \( i \) of the \( k \)-th mask matrix. The compressed matrix \( C \) also has the same shape as the model weights, and each element \( C[i] \) is an integer, calculated as follows:

\[
C[i] = \sum_{k=0}^{T-1} M_k[i] \cdot 2^k
\]

In this, \( C[i] \) stores the value of each \( M_k[i] \) using binary positional weights \( 2^k \). This method compresses the \( T \) mask values at each position \( i \) into a single integer \( C[i] \). When \( T \leq 32 \), each \( C[i] \) can be fully represented by a 32-bit integer. This binary method allows compressing 32 Mask matrices into one model size, showing high-efficiency compression performance in practical applications.

\section{Experiments}
\subsection{Implementation Detail}

We used the official ResNet50 as the backbone with input images resized to \(224 \times 224\). For both task incremental and domain incremental settings, we employed the Adam optimizer with learning rates set to 5e-5 and 3e-4, respectively. We utilized the ReduceLROnPlateau optimization technique to dynamically adjust the learning rate during training. This algorithm automatically decreases the learning rate when validation accuracy plateaus, until it reaches 1e-5, with a factor of 0.3 and a patience of 5 epochs.

Under the task incremental setting, we used a batch size of 32, trained for 100 epochs, with a network sparsity of 0.4. In the domain incremental setting, we employed a batch size of 48, trained for 80 epochs, with a network sparsity of 0.85.

\subsection{Result}

\begin{table}[!ht]
\centering
\caption{We report the average accuracy of our method under task incremental and domain incremental settings during the preliminary stages.}
\begin{tabular}{ccc}
\toprule[1.5pt]
\# & \textbf{Method}  &  \textbf{Average Accuracy} \\ 
\midrule 
1 & WSN based Baseline & 69.27 \\
2 & \textbf{+} Freeze batch normalization layers & 73.15 \\
3 & \textbf{+} Infer task ID & 76.27 \\
4 & \textbf{+} Gradient Supplementation & 78.13 \\

\bottomrule[1.5pt]
\label{table-1}
\end{tabular}
\end{table}
With Original WSN
As reported in table \ref{table-1}, with WSN base method, the average accuracy reached 69.27 score. By freezing the batch normalization layers, the average accuracy reached 73.15. By incorporating inference task IDs, the accuracy rose to 76.27, and through gradient supplementation, we ultimately achieved an average accuracy of 78.13 in the preliminary stage.

\section{Conclusion} This report summarizes our solution for the sequential task continual learning track of the 2nd Greater Bay Area International Algorithm Competition. Our approach, based on partitioning subnetworks using winning subnetworks, integrates strategies such as freezing batch normalization layers and classification layers, along with inference task identity strategy, gradient supplementation, dynamic parameter adjustment, and matrix compression strategies tailored for domain incremental learning. Our method effectively mitigates catastrophic forgetting in continual learning without requiring data replay or significant expansion of network parameters. It enhances learning accuracy for new tasks while maintaining excellent storage and inference performance. Our approach demonstrates outstanding performance in both task incremental and domain incremental learning scenarios. The final competition results show the effectiveness of our solution.

{\small
\bibliographystyle{ieee_fullname}
\bibliography{main}

\begin{thebibliography}{10}\itemsep=-1pt

\bibitem{der}
Pietro Buzzega, Matteo Boschini, Angelo Porrello, Davide Abati, and Simone Calderara.
\newblock Dark experience for general continual learning: a strong, simple baseline.
\newblock {\em Advances in neural information processing systems}, 33:15920--15930, 2020.

\bibitem{Co2L}
Hyuntak Cha, Jaeho Lee, and Jinwoo Shin.
\newblock Co2l: Contrastive continual learning.
\newblock In {\em Proceedings of the IEEE/CVF International conference on computer vision}, pages 9516--9525, 2021.

\bibitem{OGD}
Mehrdad Farajtabar, Navid Azizan, Alex Mott, and Ang Li.
\newblock Orthogonal gradient descent for continual learning.
\newblock In {\em International Conference on Artificial Intelligence and Statistics}, pages 3762--3773. PMLR, 2020.

\bibitem{CaSSLe}
Enrico Fini, Victor G~Turrisi Da~Costa, Xavier Alameda-Pineda, Elisa Ricci, Karteek Alahari, and Julien Mairal.
\newblock Self-supervised models are continual learners.
\newblock In {\em Proceedings of the IEEE/CVF Conference on Computer Vision and Pattern Recognition}, pages 9621--9630, 2022.

\bibitem{wsn}
Haeyong Kang, Rusty John~Lloyd Mina, Sultan Rizky~Hikmawan Madjid, Jaehong Yoon, Mark Hasegawa-Johnson, Sung~Ju Hwang, and Chang~D Yoo.
\newblock Forget-free continual learning with winning subnetworks.
\newblock In {\em International Conference on Machine Learning}, pages 10734--10750. PMLR, 2022.

\bibitem{ewc}
James Kirkpatrick, Razvan Pascanu, Neil Rabinowitz, Joel Veness, Guillaume Desjardins, Andrei~A Rusu, Kieran Milan, John Quan, Tiago Ramalho, Agnieszka Grabska-Barwinska, et~al.
\newblock Overcoming catastrophic forgetting in neural networks.
\newblock {\em Proceedings of the national academy of sciences}, 114(13):3521--3526, 2017.

\bibitem{connector}
Guoliang Lin, Hanlu Chu, and Hanjiang Lai.
\newblock Towards better plasticity-stability trade-off in incremental learning: A simple linear connector.
\newblock In {\em Proceedings of the IEEE/CVF Conference on Computer Vision and Pattern Recognition}, pages 89--98, 2022.

\bibitem{hat}
Joan Serra, Didac Suris, Marius Miron, and Alexandros Karatzoglou.
\newblock Overcoming catastrophic forgetting with hard attention to the task.
\newblock In {\em International conference on machine learning}, pages 4548--4557. PMLR, 2018.

\bibitem{afec}
Liyuan Wang, Mingtian Zhang, Zhongfan Jia, Qian Li, Chenglong Bao, Kaisheng Ma, Jun Zhu, and Yi Zhong.
\newblock Afec: Active forgetting of negative transfer in continual learning.
\newblock {\em Advances in Neural Information Processing Systems}, 34:22379--22391, 2021.

\bibitem{wang2024comprehensive}
Liyuan Wang, Xingxing Zhang, Hang Su, and Jun Zhu.
\newblock A comprehensive survey of continual learning: Theory, method and application.
\newblock {\em IEEE Transactions on Pattern Analysis and Machine Intelligence}, 2024.

\bibitem{Adam-NSCL}
Shipeng Wang, Xiaorong Li, Jian Sun, and Zongben Xu.
\newblock Training networks in null space of feature covariance for continual learning.
\newblock In {\em Proceedings of the IEEE/CVF conference on Computer Vision and Pattern Recognition}, pages 184--193, 2021.

\bibitem{replay_2}
Zhen Wang, Liu Liu, Yiqun Duan, and Dacheng Tao.
\newblock Continual learning through retrieval and imagination.
\newblock In {\em Proceedings of the AAAI Conference on Artificial Intelligence}, volume~36, pages 8594--8602, 2022.

\bibitem{yang2024alignment}
Yang Yang, Jinyi Guo, Guangyu Li, Lanyu Li, Wenjie Li, and Jian Yang.
\newblock Alignment efficient image-sentence retrieval considering transferable cross-modal representation learning.
\newblock {\em Frontiers of Computer Science}, 18(1):181335, 2024.

\bibitem{yang2024robust}
Yang Yang, Nan Jiang, Yi Xu, and De-Chuan Zhan.
\newblock Robust semi-supervised learning by wisely leveraging open-set data.
\newblock {\em IEEE Transactions on Pattern Analysis and Machine Intelligence}, 2024.

\bibitem{yang2021learning}
Yang Yang, Zhen-Qiang Sun, Hengshu Zhu, Yanjie Fu, Yuanchun Zhou, Hui Xiong, and Jian Yang.
\newblock Learning adaptive embedding considering incremental class.
\newblock {\em IEEE Transactions on Knowledge and Data Engineering}, 35(3):2736--2749, 2021.

\bibitem{yang2021corporate}
Yang Yang, Jia-Qi Yang, Ran Bao, De-Chuan Zhan, Hengshu Zhu, Xiao-Ru Gao, Hui Xiong, and Jian Yang.
\newblock Corporate relative valuation using heterogeneous multi-modal graph neural network.
\newblock {\em IEEE Transactions on Knowledge and Data Engineering}, 35(1):211--224, 2021.

\bibitem{yang2023contextualized}
Yang Yang, Chubing Zhang, Xin Song, Zheng Dong, Hengshu Zhu, and Wenjie Li.
\newblock Contextualized knowledge graph embedding for explainable talent training course recommendation.
\newblock {\em ACM Transactions on Information Systems}, 42(2):1--27, 2023.

\bibitem{yang2024not}
Yang Yang, Yuxuan Zhang, Xin Song, and Yi Xu.
\newblock Not all out-of-distribution data are harmful to open-set active learning.
\newblock {\em Advances in Neural Information Processing Systems}, 36, 2024.

\bibitem{yang2019adaptive}
Yang Yang, Da-Wei Zhou, De-Chuan Zhan, Hui Xiong, and Yuan Jiang.
\newblock Adaptive deep models for incremental learning: Considering capacity scalability and sustainability.
\newblock In {\em Proceedings of the 25th ACM SIGKDD International Conference on Knowledge Discovery \& Data Mining}, pages 74--82, 2019.

\bibitem{yang2021cost}
Yang Yang, Da-Wei Zhou, De-Chuan Zhan, Hui Xiong, Yuan Jiang, and Jian Yang.
\newblock Cost-effective incremental deep model: Matching model capacity with the least sampling.
\newblock {\em IEEE Transactions on Knowledge and Data Engineering}, 35(4):3575--3588, 2021.

\bibitem{zhang2021harmi}
Xiao Zhang, Hongzheng Yu, Yang Yang, Jingjing Gu, Yujun Li, Fuzhen Zhuang, Dongxiao Yu, and Zhaochun Ren.
\newblock Harmi: Human activity recognition via multi-modality incremental learning.
\newblock {\em IEEE Journal of Biomedical and Health Informatics}, 26(3):939--951, 2021.

\bibitem{zhou2021learning}
Da-Wei Zhou, Yang Yang, and De-Chuan Zhan.
\newblock Learning to classify with incremental new class.
\newblock {\em IEEE Transactions on Neural Networks and Learning Systems}, 33(6):2429--2443, 2021.

\end{thebibliography}
}

\end{document}